\pdfoutput=1
\documentclass[11pt]{article}

\usepackage[preprint]{coling}

\usepackage{times}
\usepackage{latexsym}
\usepackage{algorithm}
\usepackage{algpseudocode}
\usepackage{tabularx}
\usepackage{amsmath}
\usepackage{caption}
\usepackage{array}
\usepackage{booktabs}
\usepackage{float}
\usepackage{multirow}
\usepackage{graphicx}
\usepackage{ragged2e}

\usepackage[T1]{fontenc}
\usepackage[utf8]{inputenc}

\usepackage{microtype}

\usepackage{inconsolata}

\usepackage{graphicx}

\title{LowCLIP: Adapting the CLIP Model Architecture for Low-Resource Languages in Multimodal Image Retrieval Task}

\author{
 \textbf{Ali Asgarov\textsuperscript{1}},
 \textbf{Samir Rustamov\textsuperscript{1,2}},
\\
\\
 \textsuperscript{1}George Washington University,
 \textsuperscript{2}ADA University
\\
}

\begin{document}
\maketitle
\begin{abstract}
%% Text of abstract
This research explores the development of multimodal vision-language models for image retrieval in low-resource languages, specifically Azerbaijani. Existing vision-language models primarily support high-resource languages and fine-tuning them remains computationally demanding. To address challenges in vision-language retrieval for low-resource languages, we integrated the CLIP model architecture and employed several techniques to balance computational efficiency with performance. These techniques include synthetic data generation through machine translation, image augmentation, and further training the attention mechanisms of transformer-based models with domain-specific data. We integrated Multilingual BERT as a text encoder with image encoders like ResNet50, EfficientNet0, Vision Transformer (ViT), and Tiny Swin Transformer. Our study found that models like EfficientNet0 and Tiny Swin Transformer perform best on the datasets they were trained on, such as COCO, Flickr30k, and Flickr8k. Augmentation techniques boosted EfficientNet0’s MAP on Flickr30k from 0.84 to 0.87 and ResNet50’s MAP on MSCOCO from 0.70 to 0.80, contributing to a new state-of-the-art (SOTA) in vision-language retrieval. We share our configurations and results to support further research. Code and pre-trained models are available at \url{https://github.com/aliasgerovs/azclip}.
\end{abstract}

\section{Introduction}

The digital world is overflowing with vast amounts of information. Text, images, and videos are produced at an exceptionally high rate, and traditional search systems, designed for textual queries, struggle to keep pace. Keyword-based searches often yield extensive results that fail to capture the user's intent or the richness of multimedia data, creating barriers to accessing the needed information. Ideally, information retrieval systems should allow users to find what they need, regardless of their native language or preferred mode of interaction. 
\begin{figure} [!htbp] \centering \includegraphics[width=1\linewidth]{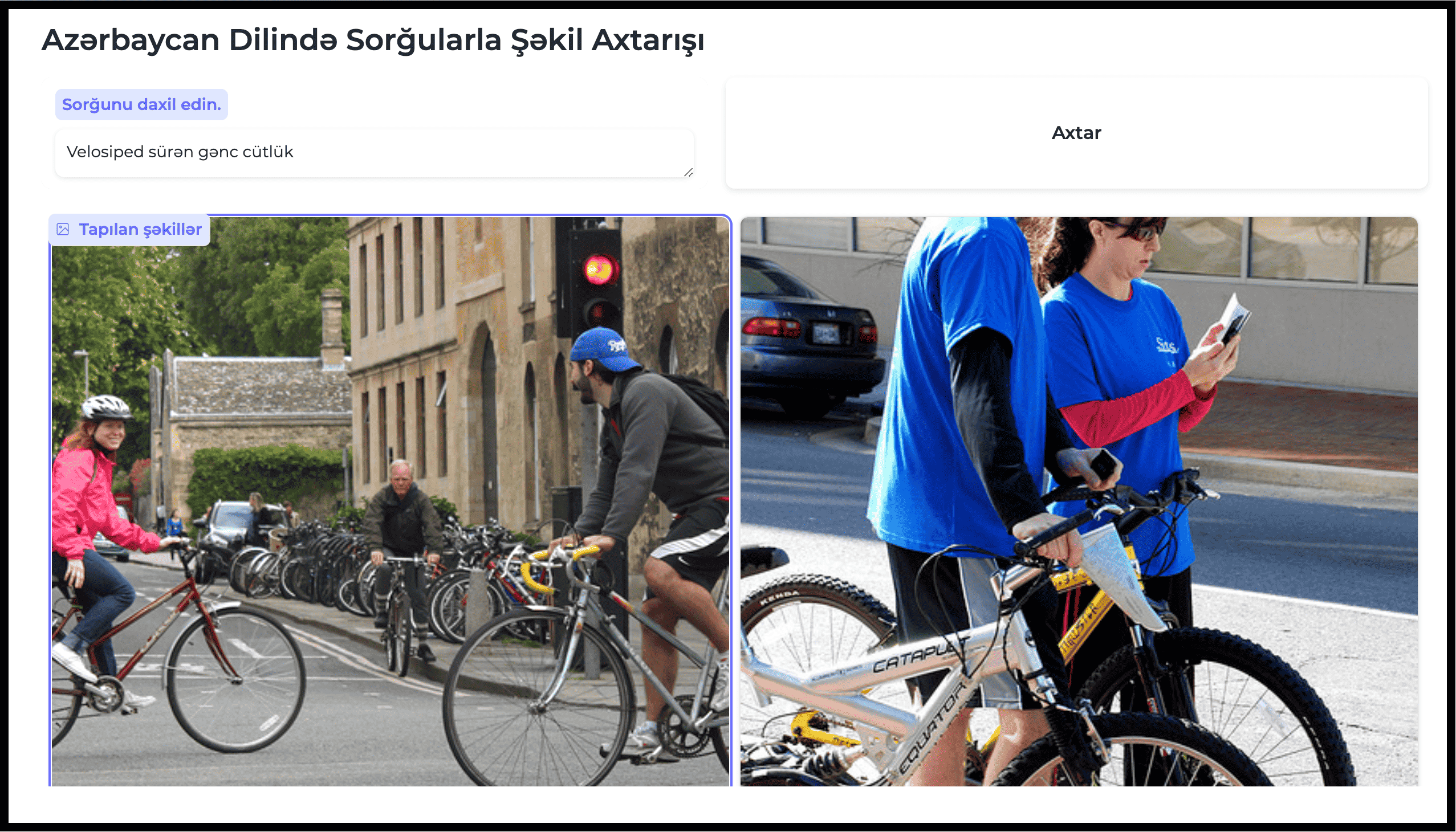} \caption{\textbf{Model Inference Example 1}. The query posed is "Young couple cycling" and the model returned relevant images from the database for the query.} \label{fig
} \end{figure}
This is where multimodal image retrieval becomes essential, as it enables searches using not only text but also pictures, spoken words, or a combination of different input modalities. This approach significantly improves search capabilities, making information more accessible to people, regardless of the language they speak or how they prefer to search. For instance, in an image-to-image search, one could point a camera at a building to search for its architectural style or use a drawing to find clothes online. These examples illustrate how multimodal data search can help people search more efficiently. However, a significant challenge exists, most multimodal data retrieval systems depend on large, complex models trained on vast datasets. These models are resource-intensive and require extensive training data in specific languages, creating a challenge for languages with limited digital resources. This research aims to close that gap, making the benefits of such models accessible to all. By making these systems more memory and computation-efficient and leveraging the vast amount of available image data in all languages, we can unlock the potential of multimodal retrieval for all languages. Most recent multimodal systems suffer from deficiencies in low-resource linguistic settings where they are less efficient in scalability with high volume and high dimensional data. The imbalance in data types' availability complicates the combination of such data for researching or training multimodal systems in languages with limited digital products. Additionally, the computational requirements of advanced models like CLIP are very high and cannot be adjusted for limited computational resources. Consequently, systems built on CLIP or similar models with billions of parameters are not widely applicable for low-resource languages. The primary objective of this research is to develop a multimodal vision-language retrieval system adapted for the Azerbaijani language, balancing computational efficiency and scalability in low-resource settings. An important part of this research is to analyze ways of training the model's performance across different domains, focusing on the constrained availability of data. The aim is to create a model that performs well within the Azerbaijani language and represents a scalable skeleton for other similar low-resource languages, such as Kazakh or Uzbek. This approach aims to expand the use of powerful AI technologies into various linguistically diverse and low-resource settings, making computationally challenging problems accessible to a broader audience.

Our main contributions are:
\begin{itemize}
\item Development and extensive validation of a multimodal vision-language retrieval model specifically for the Azerbaijani language, creating a custom image retrieval model that can effectively perform in a low-resourced linguistic environment.
\item Contribution to the field of computational efficiency of model designs. The model can be easily replicated for other low-resource languages, and its high computational efficiency decreases operational demands and allows for operational deployment.
\item Comparative analysis of visual encoder and text decoder models and their performance on in-domain and out-of-domain data to estimate the process of generalization and scaling, expanding on the concept by assessing how they adapt to new, unseen environments.
\end{itemize}

\section{Literature Review}
\subsection{Introduction to Multimodal Retrieval}

Multimodal retrieval systems have become an important study area in artificial intelligence as they are concerned with the analysis of data across multiple modalities, such as vision and text or speech etc. One of the foundational contributions to this field is the Deep Visual-Semantic Embedding model by Frome et al  \cite{10.5555/2999792.2999849}. This model is crucial, since it combines visual and textual data within a single representation space and it has inspired many upcoming works in this area. Then, Radford developed CLIP \cite{radford2021learning} – Contrastive Language Image Pretraining, that translates into using descriptions in natural language to recognize visual concepts. Later, newer architectures were developed, which enabled better interaction of textual and visual data. Both of VisualBERT \cite{li2019visualbert} and ViLBERT \cite{lu2019vilbert} use transformer design as a part of their architecture to receive input in both visual and text views. LXMERT \cite{tan2019lxmert} is another example of the recent models, that suited to the more general-purpose multimodal tasks that involve better understanding of images and texts, such as visual question answering. One of the most recent works in this direction was the integration of multimodal data processing with object detection by use of MDETR \cite{kamath2021mdetr}. It has very unique architecture, as it uses the attention mechanism in such an architecture to take control of the object detection process based on the textual description provided. Overall, the systems mentioned follow the trend of integrating visual and textual data that began with the very early models like DeViSE \cite{10.5555/2999792.2999849} and CLIP \cite{radford2021learning} and has expanded to include increasingly diverse tasks, from object recognition to translation.

\subsection{History of Image Retrieval Techniques}

Recent advancements in image retrieval have been significantly influenced by the integration of deep learning techniques, particularly with the shift from keyword-based methods to more advanced contextual analysis. Early approaches, such as Swain and Ballard's color indexing, represented a move away from external keyword reliance towards leveraging intrinsic image features. Convolutional Neural Networks (CNNs) enabled the extraction of complex image features, improving the retrieval process by recognizing higher-level content within images. Multimodal learning, as explored by Srivastava and Salakhutdinov \cite{srivastava2012multimodal}, further developed the field by using both text and image data to learn deep representations. Models like AlexNet and more recent transformer-based architectures have refined the extraction and matching of image features with visual vocabularies, significantly enhancing task efficiency and enabling real-time processing, as shown by Joulin et al. (2016) \cite{joulin2016learning} and Radford et al. (2021) \cite{radford2021learning}.

\subsection{Dataset Development and Challenges}

The creation of large-scale datasets such as COCO, Flickr 8k/30k, and ROCO has been essential for training and evaluating multimodal models, despite some criticisms regarding representativeness and cultural bias \cite{sharma2018conceptual}. COCO, with its extensive collection of images and object categories, has played a key role in object detection tasks \cite{lin2014microsoft}, while specialized datasets like ROCO have addressed specific needs in medical imaging \cite{pino2019roco}. However, challenges remain, particularly in low-resource language contexts where data scarcity and computational limitations reduce the effectiveness of multimodal models \cite{agic2020data, lewis2020pretraining}. Recent developments in image and text encoding methods, including ResNet \cite{he2016deep}, Vision Transformers \cite{dosovitskiy2021an}, and BERT \cite{devlin2018bert}, have advanced feature extraction and understanding. However, these advancements also highlight the need for efficient, adaptable models in environments with limited resources. Synthetic data generation, as demonstrated by Varol et al. (2017) \cite{varol2017learning}, offers a valuable approach to addressing these challenges by providing high-quality training data where real data is scarce or sensitive.

\section{Research Methodology}
\subsection{Method}

The core of the system is based on two main components: an image encoder and a text encoder, which together process and compare visual and textual data. The image encoder uses a pretrained network (for example, ResNet50), which is adapted to ignore the final classification layer and instead focus on extracting feature representations of images. The text encoder is built around a BERT model that has been pretrained to understand multiple languages, making it suitable for processing text in various languages (BertModel from Hugging Face). To establish the connection between the image and text representations, we use a projection head on both encoders’ outputs. Accordingly, the component’s focus is to decrease the dimensionality of the outputs to the similar size. Meanwhile, the technique that allows to carry out the actual comparison of image and text features is contrastive loss, which also assists the model in understanding which images and texts it has been learned to relate.

\begin{algorithm}
\caption{Training a CLIP Model}
\begin{algorithmic}[1]
\State \textbf{Input:} Training and validation datasets
\State Initialize tokenizer, image and text encoders, projection heads
\State Load datasets and data loaders

\For{\texttt{epoch = 1, 2, ..., NumEpochs}}
    \For{\texttt{images, input\_ids, attention\_mask in train\_loader}}
        \State Move images, input\_ids, attention\_mask to device
        \State Zero the gradients
        \State Compute image and text embeddings
        \State Calculate contrastive loss
        \State Backpropagate error and update model weights
        \State Update learning rate
        \State Log loss and other metrics
    \EndFor
    \If{\texttt{current loss $<$ best loss}}
        \State Update best loss
        \State Save model checkpoint
    \EndIf
\EndFor
\end{algorithmic}
\end{algorithm}

\subsection{Dataset and Preprocessing}

This research is based on multiple existing and recently collected datasets, enriched with Azerbaijani translations to facilitate more effective training in a less studied language in machine learning. A critical dataset is the MS COCO dataset, used for detailed image comprehension and object recognition. It contains a vast amount of visual information, with thorough descriptions and detailed segmentation, making it valuable for training models to understand complex scenes in natural settings (Lin et al., 2014) \cite{lin2014microsoft}. Another one is the Flickr30k dataset that comprises 31,000 images from Flickr, each described by five different captions detailing the scenes, objects, and activities depicted, enhancing the model’s ability to interpret and generate nuanced text about images (Young et al., 2014) \cite{young2014image}. The ROCO (Radiology Objects in Context) dataset includes medical imaging data, such as radiological scans paired with descriptive texts, extending the model's capability to process and understand medical imagery (Pelka et al., 2018) \cite{pelka2018roco}.

\begin{figure}  % The [H] specifier forces the figure to be placed exactly here
  \centering
  \includegraphics[width=0.8\linewidth]{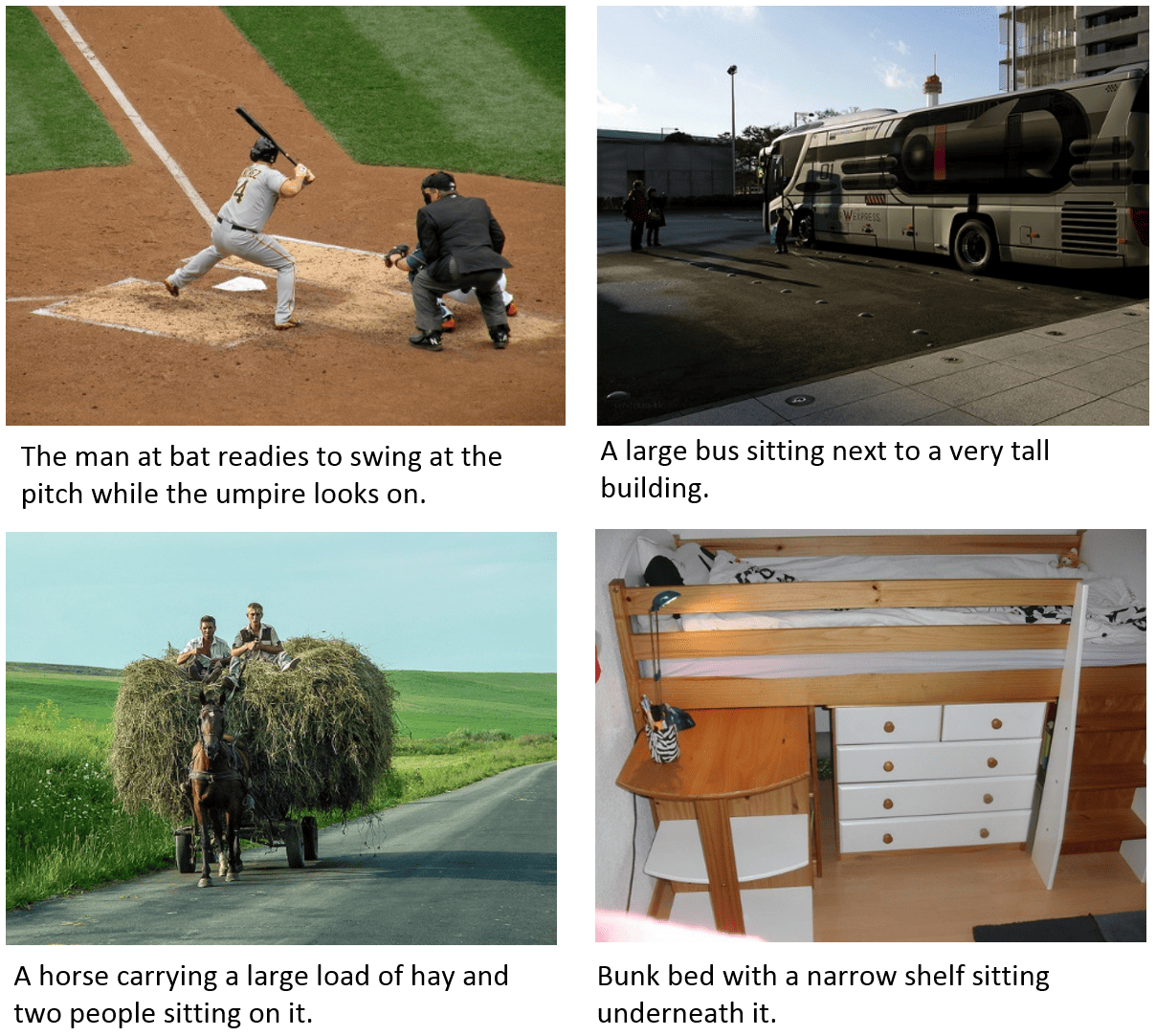}
  \caption{\textbf{ COCO Dataset}. Example images and captions from the Microsoft COCO Caption dataset \cite{lin2014microsoft}.}\label{fig:teaser}
\end{figure}

\noindent
\subsubsection{Generated Dataset for Azerbaijani Language}
\begin{figure}[!htbp]
    \centering
    \includegraphics[width=0.7\linewidth]{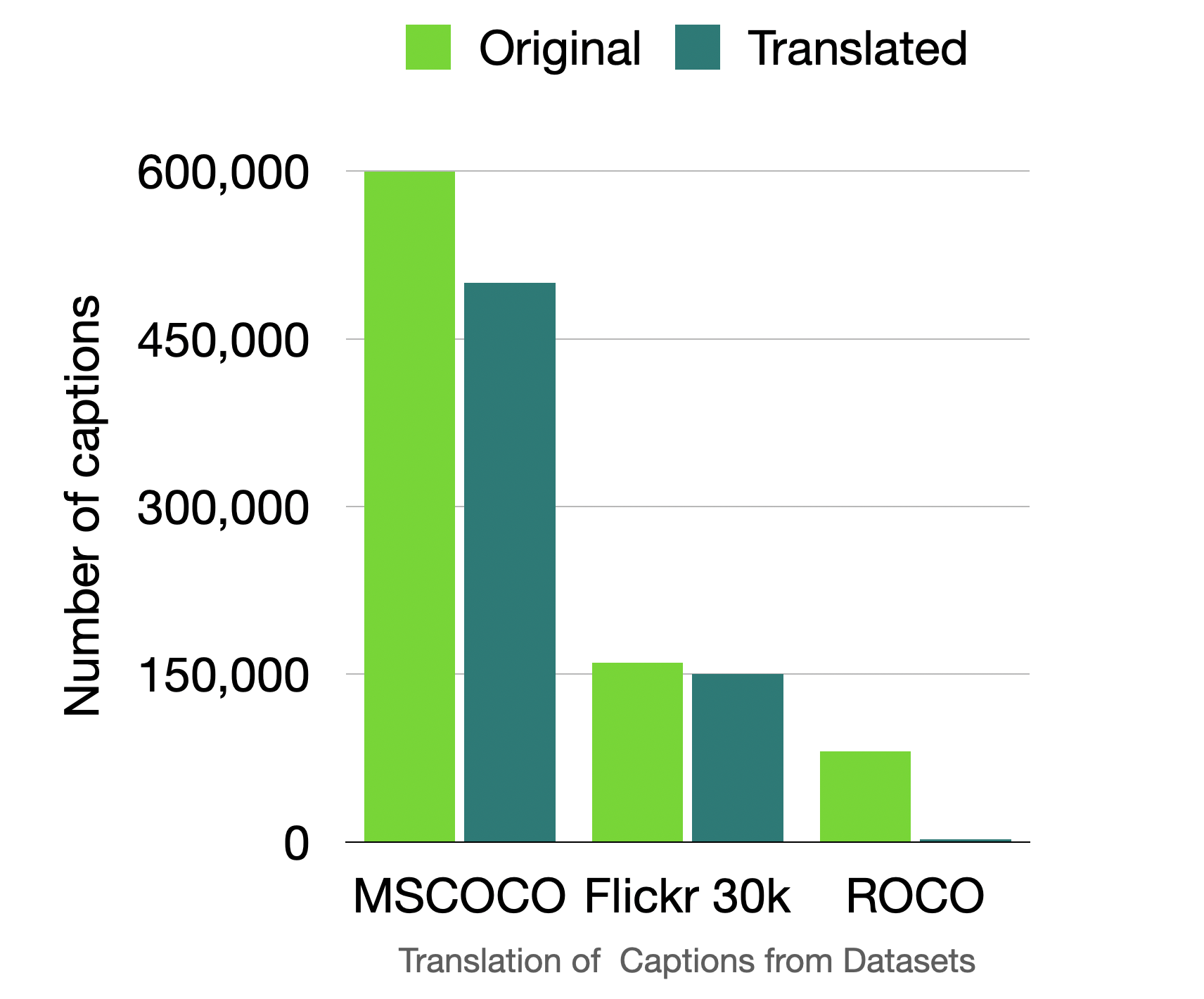}
    \caption{\textbf{Translation of  Captions from Datasets}. Translation of  Captions from COCO, Flickr30k and ROCO datasets.}
    \label{fig:translation}
\end{figure}

To support the Azerbaijani language, as it is low resource language, this study incorporates image-text pairs in Azerbaijani. What we have done is, we have translated more than half a million captions from COCO dataset from English to Azerbaijani, and 90\% of Flickr30k captions (Figure \ref{fig:translation}). 

\noindent
The datasets underwent several preprocessing steps to ensure high-quality data for model training:

\noindent
\textbf{Data Cleaning:} The main steps in preparing datasets involved the exclusion and removal of corrupt or irrelevant captions from ther dataset. There were some images that did not have 5 captions available for the, we needed to remove those samples since it creates bias in the dataset. We have also used other types of data cleaning techniques, to be sure that we do not have special characters and also noise in our dataset. 

\noindent
\textbf{Text Translation:} Texts associated with images from the Microsoft COCO dataset, with over 500,000 captions, more than 40,000 captions from the Flickr dataset, and over 2,500 captions from the ROCO dataset, were translated into Azerbaijani to build a dataset that supports this language, broadening the model’s usability.

\subsection{Image Encoders}

The architecture depicted in Figure \ref{fig:transformerr} efficiently processes both visual and textual data, enhancing performance on tasks involving both vision and language, especially in low-resource settings due to reduced data and computational needs. By integrating image encodings and text embeddings, and using smaller models, it avoids extensive fine-tuning while enhancing encoding capabilities through image augmentation and domain-specific text data, improving the attention mechanism within a multi-language BERT model. The image encoder utilizes pre-trained CNN models with the final layer modified to an identity mapping for rich image representation. Various CNN architectures, such as ResNet-50, EfficientNet-B0, Vision Transformer (ViT), Swin Transformer, and ConvNeXt, were tested for their feature extraction capabilities in multimodal retrieval tasks. Each model was evaluated on metrics such as top-1 and top-5 accuracy on ImageNet-1K, number of parameters, GFLOPS, and model file size, providing a basis for selecting the appropriate architecture based on specific application requirements.

\begin{table}[htbp]
\centering
\footnotesize
\setlength{\tabcolsep}{3pt}
\begin{tabular}{@{}lccccc@{}}
\toprule
\textbf{Model} & \textbf{Top-1} & \textbf{Top-5} & \textbf{Params} & \textbf{GFLOPs} & \textbf{Size} \\
\midrule
ResNet-50   & 76.13 & 92.86 & 25.6M & 4.09 & 97.8 \\
EffNet-B0   & 77.69 & 93.53 & 5.29M & 0.39 & 20.5 \\
ViT Base    & 81.07 & 95.32 & 86.6M & 17.56 & 330.3 \\
Swin-Tiny   & 81.47 & 95.78 & 28.3M & 4.49 & 108.2 \\
ConvNeXt T  & 82.52 & 96.15 & 28.6M & 4.46 & 109.1 \\
\bottomrule
\end{tabular}
\caption{Specs of DL Models on ImageNet-1K (1K categories). Top-1 and Top-5 are accuracy percentages. Size is in MB. Minimum input size: ResNet-50 and EffNet-B0: 1$\times$1, ViT Base and Swin-Tiny: 224$\times$224, ConvNeXt Tiny: 32$\times$32.}
\label{table:compact_model_specs}
\end{table}

\subsection{Text Encoder}
\begin{figure}[!htbp]
    \centering
    \includegraphics[width=0.9\linewidth]{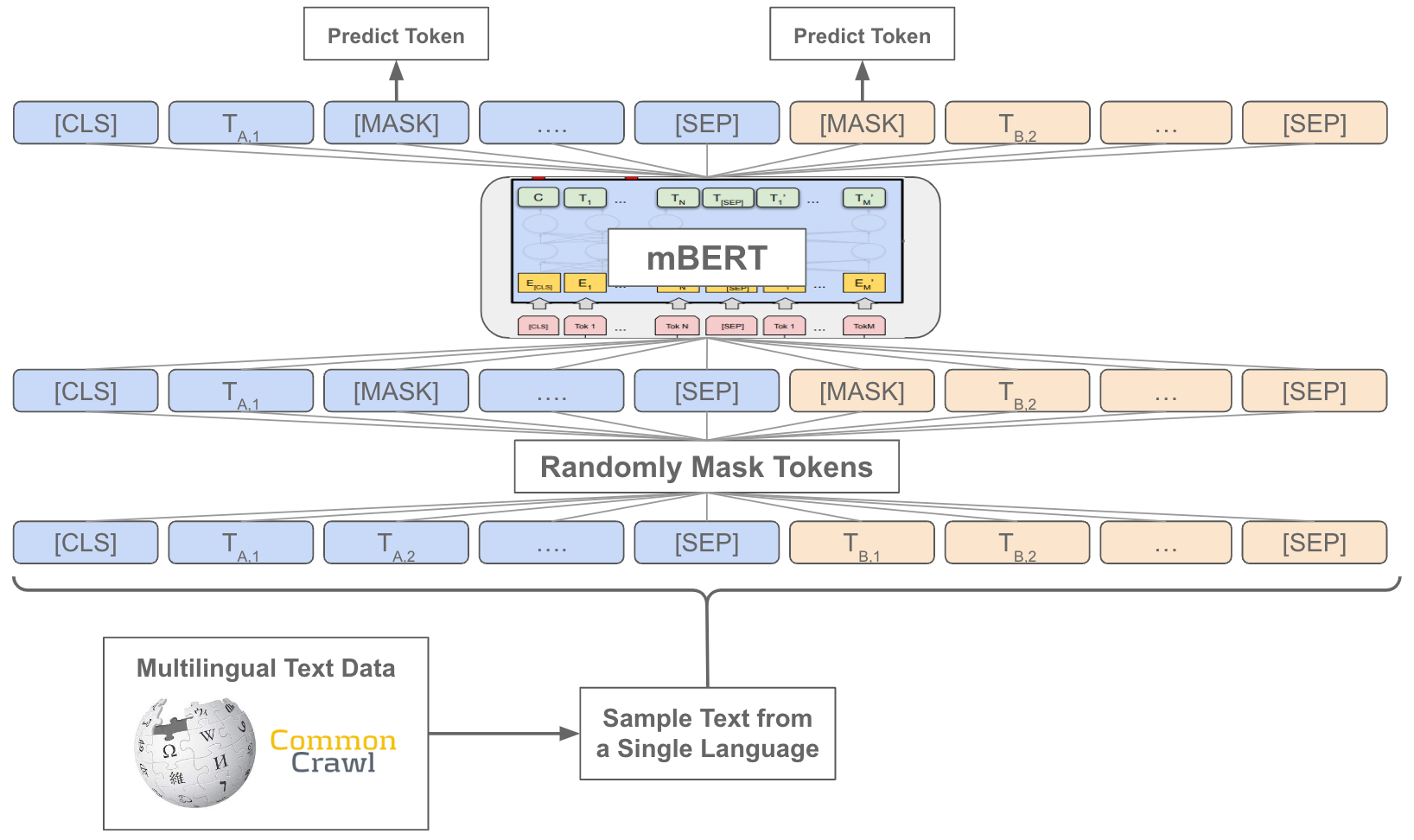}
    \caption{\textbf{Multilingual BERT training approach using masked language modeling}. \cite{devlin2019bert}}
    \label{fig:mbert}
\end{figure}

We employed a multilingual BERT \cite{devlin2019bert} as our text encoder, optimized for low-resource languages with tokenization support for over 104 languages, ideal for applications requiring diverse linguistic support (see Figure \ref{fig:mbert}).

\subsubsection{Projection Head}
\begin{figure}[!htbp]
    \centering
    \includegraphics[width=0.8\linewidth]{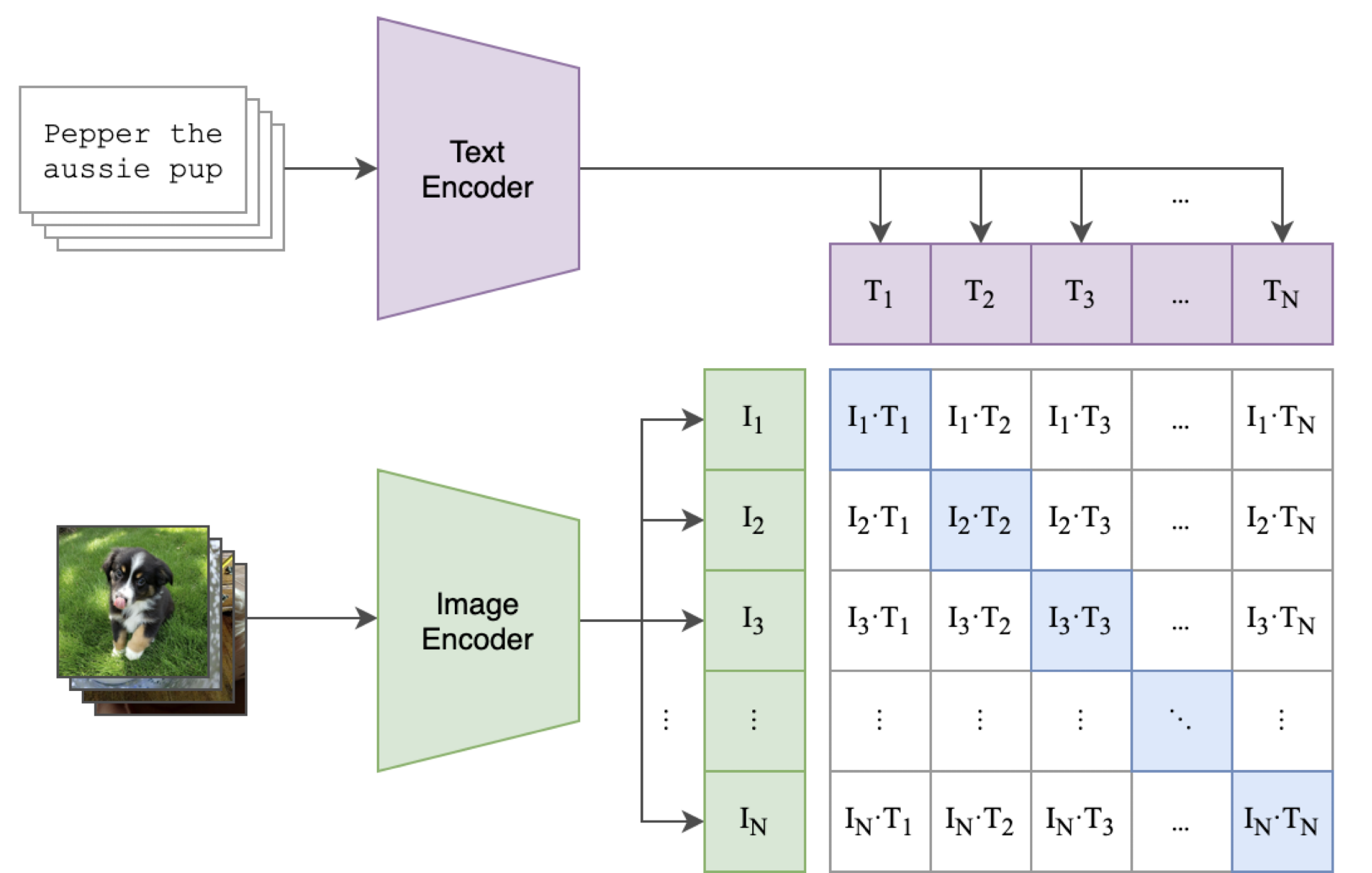}
    \caption{\textbf{Projection Head:} Framework for computing compatibility scores for image-text pairs. \cite{radford2021learning}}
    \label{fig:transformerr}
\end{figure}

The image and text encoders feature a projection head to map high-dimensional features into a shared embedding space, enhancing contrastive learning efficiency (Figure \ref{fig:transformerr}).

\subsubsection{Contrastive Learning Loss}
The model utilizes a contrastive learning loss, structured to maximize the similarity of correct image-text pairs while minimizing the similarity of incorrect ones, incorporating a margin-based penalty for better discrimination. (Equations (\ref{eq:total_loss}, \ref{eq:l1}, \ref{eq:l2}, \ref{eq:mij}).

\begin{alignat}{2}
L &= -\sum_{i=1}^{N} \left( L_1 + L_2 \right) \label{eq:total_loss} \\[1.2em]
L_1 &= \log \frac{e^{s_{ii}}}{\sum_j e^{s_{ij}}} \label{eq:l1} \\[1.2em]
L_2 &= \lambda \cdot \left[ \max(0, m - M_{ij}) \right]^2 \label{eq:l2} \\[1.2em]
M_{ij} &= \min_{j \neq i} s_{ij} \label{eq:mij}
\end{alignat}

\subsubsection{Enhancing Feature Extraction}

Adaptive Feature Learning and Multi-Resolution Feature Extraction techniques were introduced to optimize feature extraction from images, adjusting the network’s response based on the image content and processing images at various scales. Retraining Multilingual BERT with domain-specific data from Wikipedia and employing Hierarchical Attention Networks improve the model’s text feature extraction capabilities, allowing it to handle complex multi-modal inputs more efficiently. By integrating these techniques for both image and text feature extraction, the system achieves a more robust and adaptive feature extraction process, enhancing its overall performance in multi-modal tasks.

\subsubsection{Training Procedure}
\begin{figure}[!htbp]
    \centering
    \begin{minipage}{0.48\textwidth}
        \centering
        \includegraphics[width=\linewidth]{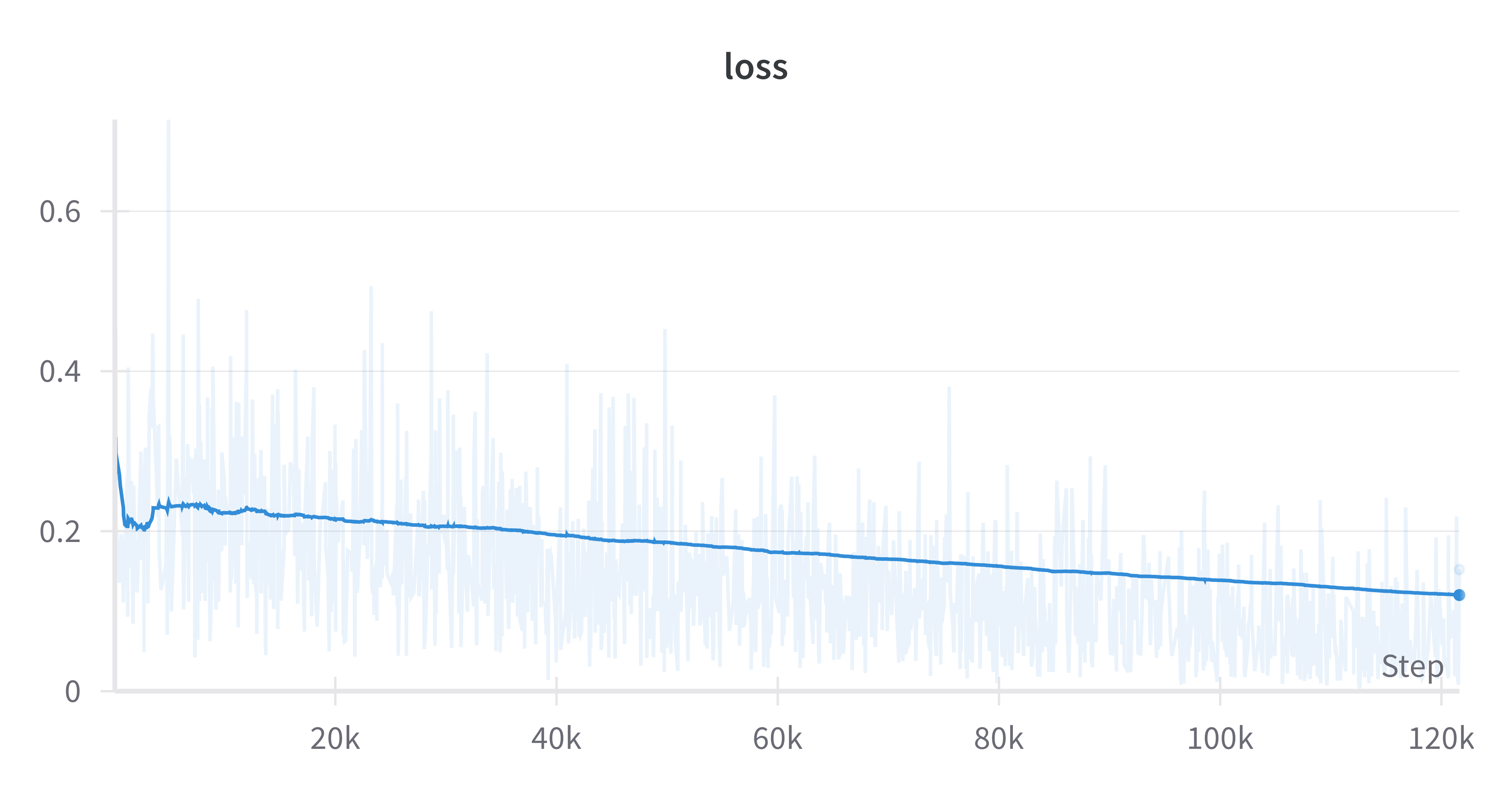}
        \caption{\textbf{Base Model (AzClip) Training Loss}}
        \label{fig:base_model_loss}
    \end{minipage}\hfill
    \begin{minipage}{0.48\textwidth}
        \centering
        \includegraphics[width=\linewidth]{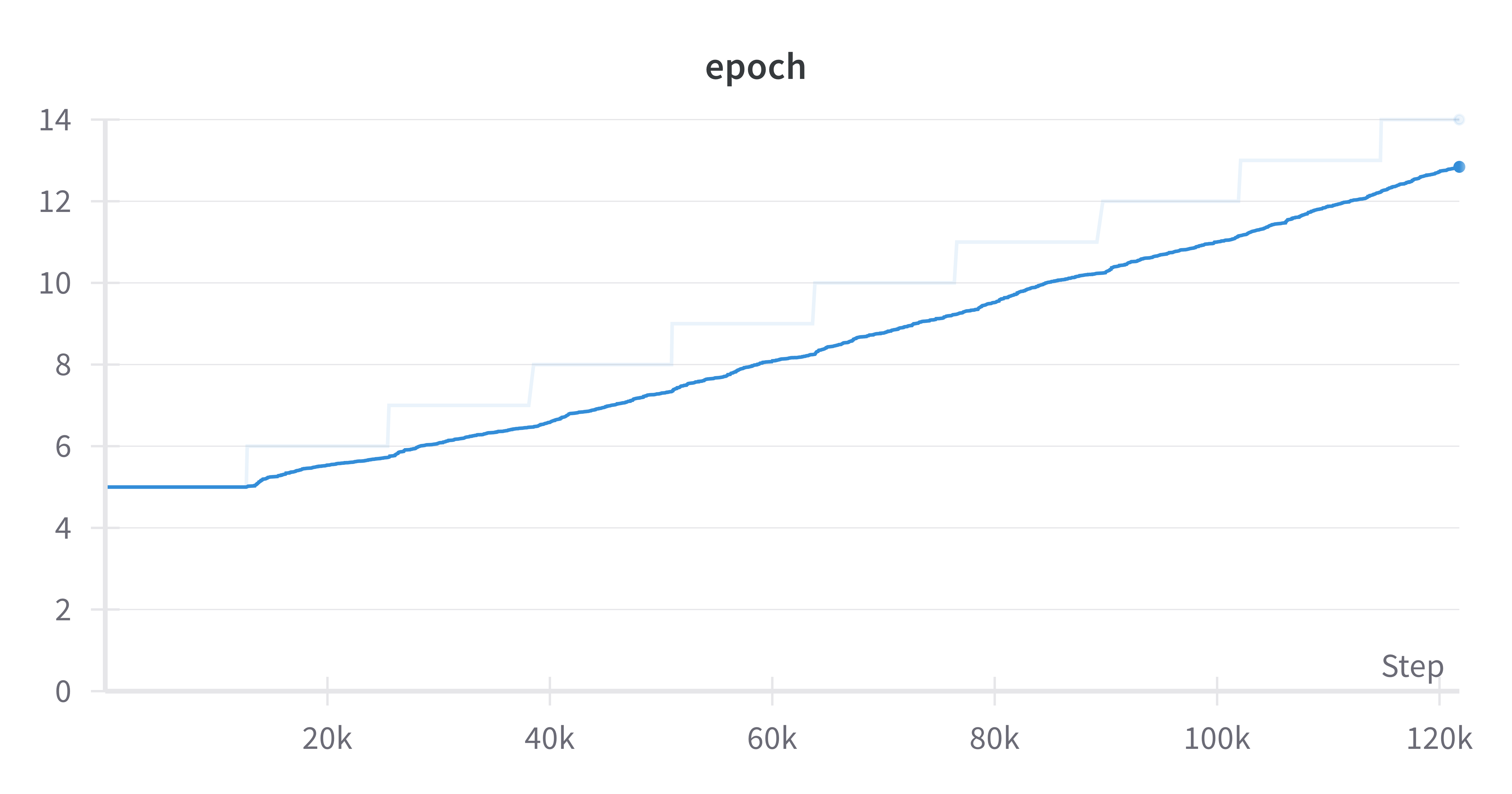}
        \caption{\textbf{Base Model (AzClip) Epochs}}
        \label{fig:base_model_epochs}
    \end{minipage}
\end{figure}

The training leverages AdamW optimizer and dynamic learning rate scheduling to optimize model performance. Regularization techniques such as dropout and data augmentation were implemented to combat overfitting (Figure \ref{fig:base_model_loss}).

\begin{table}[h!]
    \centering
    \begin{tabular}{|c|c|}
        \hline 
        \textbf{Parameter} & \textbf{Value} \\
        \hline
        Total number of parameters & 192,309,312 \\
        \hline
        Total number of trainable parameters & 192,309,312 \\
        \hline
        Total number of epochs & 7 \\
        \hline
        Used GPU & Nvidia T4 \\
        \hline
        Total number of training hours & 37 \\
        \hline
    \end{tabular}
    \caption{Summary of Base Model Training (Resnet50 + mBERT) Parameters}
    \label{tab:model_training}
\end{table}

\subsection{Image Retrieval Process}
The retrieval process begins by loading precomputed image embeddings, which are tensor representations of visual content from the MSCOCO dataset. We have pre computed the image embeddings and saved it on the local repository, this is one time task which helps to increase the speed and avoiding the waste of computation resources in each query. The trained model is then set to evaluation mode, which is disabling training specific operations such as dropout to maintain consistency during inference. Then the given input text, query is tokenized, padded, and truncated to ensure uniform length, and converted into a text embedding using a pre trained text encoder. Then the next task is projection of this embedding using a linear layer to align with the dimensions of the image embeddings, and normalizing to facilitate scale invariant comparisons. The similarity between the text and image embeddings is being calculated using the cosine similarity metric, which is mathematically represented as:

\[
\text{Cosine Sim}(\mathbf{A}, \mathbf{B}) = \frac{\sum_{i=1}^{n} A_i \times B_i}{\sqrt{\sum_{i=1}^{n} A_i^2} \times \sqrt{\sum_{i=1}^{n} B_i^2}}
\]

where:
\begin{itemize}
    \item $\mathbf{A}$ and $\mathbf{B}$ are two vectors of length $n$.
    \item $A_i$ and $B_i$ are components of vectors $\mathbf{A}$ and $\mathbf{B}$ respectively.
\end{itemize}

The indices of the top \(k\) images with the highest similarity scores are being identified. These indices are used to retrieve the corresponding image IDs from the dataset, which represent the images most similar to the given text description. The retrieval process concludes by outputting the image IDs of the images that best match the text query, demonstrating the effectiveness of the used text-to-image matching algorithm.

\subsection{Evaluation Metrics}
For the evaluation of image retrieval tasks, it is important that metrics be chosen for use that effectively capture both accuracy and relevance to the query of the retrieved images. The following metrics have been used to quantify the performance of the image retrieval systems:
\newline

\noindent
\textbf{Precision}: Precision (P) measures the proportion of retrieved images that are relevant to the query. It is calculated using the formula:

\[
\text{P} = \frac{\text{Number of relevant images retrieved}}{\text{Total number of images retrieved}}
\]

\noindent
\textbf{Recall}
Recall (R) quantifies the proportion of relevant images that are successfully retrieved. It is given by the formula:

\[
\text{R} = \frac{\text{Number of relevant images retrieved}}{\text{Total number of relevant images}}
\]

\subsubsection{Mean Average Precision (MAP)}
The mean average precision is a metric that collects average precision scores, after every relevant image retrieved, over all query images. This provides a compositional measure of how well all multiple retrievals do. The formula for MAP is:
\begin{equation}
    \text{MAP} = \frac{1}{Q} \sum_{q=1}^{Q} \left( \frac{1}{m_q} \sum_{k=1}^{m_q} \text{Precision}_k \right)
\end{equation}
where $Q$ is the number of queries, $m_q$ is the number of relevant images for query $q$, and $\text{Precision}_k$ is the precision calculated at the cutoff $k$, the point at which each relevant image is retrieved.

\subsubsection{Mean Average Recall (MAR)}
Mean Average Recall (MAR) is a comprehensive metric that averages the recall scores after each relevant image is retrieved, across all queries. It provides an aggregated measure of performance across multiple query images. The formula for MAR is:
\begin{equation}
    \text{MAR} = \frac{1}{Q} \sum_{q=1}^{Q} \left( \frac{1}{m_q} \sum_{k=1}^{m_q} \text{Recall}_k \right)
\end{equation}
where $Q$ is the number of queries, $m_q$ is the number of relevant images for query $q$, and $\text{Recall}_k$ is the recall calculated at the cutoff $k$, the point at which each relevant image is retrieved. 

\noindent
\subsubsection{Mean Average F1 Score (MAF1)}
The Mean Average F1 Score (MAF1) is a comprehensive metric that averages the F1 scores after each relevant image is retrieved, across all queries. It is particularly useful in balancing the consideration of both precision and recall, especially in scenarios with an uneven distribution of relevant and irrelevant images. The formula for MAF1 is:
\begin{equation}
    \text{MAF1} = \frac{1}{Q} \sum_{q=1}^{Q} \left( F1_q \right)
\end{equation}
where $Q$ is the number of queries, and $F1_q$ is the F1-score for query $q$, calculated as:
\begin{equation}
    F1_q = 2 \cdot \frac{\text{Precision}_q \cdot \text{Recall}_q}{\text{Precision}_q + \text{Recall}_q}
\end{equation}
Precision$_q$ and Recall$_q$ are calculated at the point of each relevant retrieval, ensuring that the metric reflects the balance of retrieval effectiveness across all queries.

\noindent
\subsubsection{Mean Average Accuracy (Top-k Accuracy)}
The Mean Average Accuracy for Top-k (where \( k \) is 1, 5, or 10) is a critical metric that measures the accuracy of a model in identifying the correct answer within the top \( k \) predictions across all queries. This metric is particularly significant in scenarios where multiple potential correct answers are acceptable, but the precise ranking of those answers is less critical. The formulas for Mean Average Top-1, Top-5, and Top-10 Accuracy is given by:
\begin{equation}
    \text{Mean Avg Top-}k\text{-Accuracy} = 
    \begin{aligned}
        &\frac{1}{Q} \sum_{q=1}^{Q} \left( \text{Accuracy}_q^{(k)} \right)
    \end{aligned}
\end{equation}

where \( Q \) is the number of queries, and \( \text{Accuracy}_q^{(k)} \) indicates whether the correct answer appears within the top \( k \) results for query \( q \), defined as:

\begin{equation}
    \text{Accuracy}_q^{(k)} = \begin{cases} 
    1 & \text{if correct answer is in top } k, \\
    0 & \text{otherwise.}
    \end{cases}
\end{equation}

This measurement underscores the model's effectiveness in presenting highly relevant results at the top of its output list, which is especially useful in search and recommendation systems where immediate relevance is paramount.

\section{Implementation}
\subsection{Experiment Design}

This section shows how we evaluated the vision-language models that we developed. We designed the evaluation to test the ability of the models to work well over different datasets. This would help us see how they will work when presented with data they were never presented with before and also how they would do on the completely new data.

\noindent
\textbf{In Domain Testing:} Evaluation of models on the dataset they were trained on, focusing on how well they perform in a familiar data environment.

\noindent
\textbf{Out Domain Testing:} Evaluation of models on datasets other than those they were trained on to assess their generalizability and robustness across different data distributions.

\noindent
Our experiments utilize a diverse array of models coupled with a Base Multilingual BERT to handle text in the Azerbaijani language:

\noindent
\textbf{Architectures:} ResNet50, EfficientNet0, Vision Transformer (ViT), Tiny Swin Transformer.

\noindent
\textbf{Datasets:} MSCOCO, Flickr8k, Flickr30k. Synthetic datasets were also generated using machine translation and augmentation techniques to simulate low-resource conditions.

\noindent
\textbf{Metrics}: , Mean Average Precision (MAP), Mean Average Recall (MAR), Mean Average F1-Score (MAF1), Top-K Accuracy Metrics (Top1, Top5, Top10)

\begin{itemize}
    \item \textbf{ResNet50 + Multilingual BERT:} Evaluated on MSCOCO with generalization to Flickr8k and Flickr30k.
    
    \item \textbf{Augmented Data with ResNet50 + Multilingual BERT:} Assessed the impact of image augmentation on Flickr30k, with generalization to MSCOCO and Flickr8k.
    
    \item \textbf{EfficientNet0 + Multilingual BERT:} Baseline performance evaluated on Flickr30k, extended to MSCOCO and Flickr8k.
    
    \item \textbf{Augmented Data with EfficientNet0 + Enhanced Multilingual BERT:} Explored the effects of image augmentation and enhanced BERT techniques on Flickr30k.
    
    \item \textbf{ViT + Multilingual BERT:} Tested on Flickr8k with generalization to Flickr30k and MSCOCO.
    
    \item \textbf{Tiny Swin Transformer + Multilingual BERT:} Evaluated on Flickr8k with cross-dataset adaptability to MSCOCO and Flickr30k.
\end{itemize}

\subsection{Controlled Variables and Their Impact on Model Performance}

In this section, we describe the variables that were different in each experiment in a systematic manner and how they have been changed with an intent and aim to study their impacts on performance. All of these variables have provided insights into optimization and scalling of the model in different settings, especially in low-resource and different domains.

\noindent
\textbf{Model Configurations}: Different configurations of text and image encoding models, like different number of layers, activation functions, and fine-tuning depths, were experimented with to find the best combinations of these features that would balance performance with computational efficiency. This included both comparisons of the standard BERT architecture against its multilingual version and experimentation with different backbone architectures for the image encoder, including ResNet-50 versus other alternatives. This was done with the view of examining various setups offering tradeoff between accuracy and computational load.

\noindent
\textbf{Data Volume and Quality}: Sensitivity to the quality and quantity of the data was assessed by training and testing the models on datasets of different sizes and conditions, on the datasets such as MSCOCO and Flickr30k, Flickr8k. The influence of data augmentation techniques on enhancing data quality and improving the performance of models was tested to get insights into how the models might perform in unexpected real-world environment datasets.

\noindent
\textbf{Computational Constraints: }Performance measurement in experiments was targeted towards real-world deployments, especially in low-resource settings. In this regard, the performance of the models under different computational constraints was important and that is why we have chosen entry-level GPUs like the Nvidia T4. The efficiency of various model configurations tested to obtain the best trade off between the accuracy obtained and the computational demand, something that is more crucial in practice due to the computational constraints.

\noindent
When these different parameters that have been changed during different experiments could be understood, that helps in building the model not just for optimal performance in ideal conditions but also ensuring its practical applicability in diverse and challenging environments and domains. For example, if there is need to deploy the model to the another domain where no or few captions available to further train the model, model itself should be much adaptable to other domains.

\begin{itemize}
    \item Richer annotations and diverse image content in datasets like MSCOCO lead to better model performance.
    \item Augmentation of data sources consistently improves model accuracy across different settings.
    \item The performance gap between in-domain and out-domain highlights the need for models to better generalize across diverse datasets.
    \item A careful balance between computational efficiency and accuracy is crucial, especially in resource-constrained environments.
\end{itemize}

\noindent
These findings emphasize the importance of dataset quality, the benefits of augmentation, and the challenges posed by model and data heterogeneity. Addressing these challenges is key to advancing the state-of-the-art in vision-language models.

\begin{table*}[t]
\captionsetup{justification=centering}
\caption{Performance Metrics for Various Models Across Different Datasets}
\label{tab:consolidated_performance}
\centering
\resizebox{\textwidth}{!}{%
\begin{tabular}{llcccccccccc}
\toprule
Model & Dataset & \multicolumn{3}{c}{MAP} & \multicolumn{3}{c}{MAR} & \multicolumn{3}{c}{MAF1} & Top1 \\
\cmidrule(lr){3-5} \cmidrule(lr){6-8} \cmidrule(lr){9-11}
& & MSCOCO & Flickr8k & Flickr30k & MSCOCO & Flickr8k & Flickr30k & MSCOCO & Flickr8k & Flickr30k & In-domain \\
\midrule
ResNet50 + MultiBERT (Base Loss) & MSCOCO & \textbf{0.70} & 0.66 & 0.63 & \textbf{0.65} & 0.61 & 0.58 & \textbf{0.67} & 0.63 & 0.60 & 0.55 \\
ResNet50 + MultiBERT & MSCOCO & \textbf{0.80} & 0.76 & 0.73 & \textbf{0.74} & 0.70 & 0.67 & \textbf{0.77} & 0.73 & 0.70 & 0.62 \\
ResNet50 + MultiBERT (Augmented) & Flickr30k & 0.82 & 0.84 & \textbf{0.82} & 0.76 & 0.78 & \textbf{0.75} & 0.79 & 0.81 & \textbf{0.78} & 0.64 \\
EfficientNet0 + MultiBERT & Flickr30k & 0.78 & 0.82 & \textbf{0.84} & 0.77 & 0.79 & \textbf{0.81} & 0.76 & 0.80 & \textbf{0.82} & 0.67 \\
EfficientNet0 + MultiBERT (Augmented) & Flickr30k & 0.81 & 0.85 & \textbf{0.87} & 0.80 & 0.82 & \textbf{0.84} & 0.79 & 0.83 & \textbf{0.85} & 0.70 \\
ViT + MultiBERT (Base Loss) & Flickr8k & 0.71 & \textbf{0.80} & 0.70 & 0.74 & \textbf{0.82} & 0.69 & 0.72 & \textbf{0.81} & 0.70 & 0.72 \\
Tiny Swin + MultiBERT (Base Loss) & Flickr8k & 0.80 & \textbf{0.84} & 0.79 & 0.75 & \textbf{0.78} & 0.72 & 0.77 & \textbf{0.81} & 0.76 & 0.75 \\
\bottomrule
\end{tabular}%
}
\caption*{\small Note: Bold values indicate in-domain performance. MultiBERT refers to Base Multilingual BERT.}
\end{table*}

\subsection{Analysis of Results}

Our investigations into the performance of various model architectures across datasets like COCO, Flickr30k, and Flickr8k reveal the complexities of model generalization in diverse data environments. We evaluated models such as ResNet50, EfficientNet0, Vision Transformer, Tiny Swin Transformer, and Base Multilingual BERT, observing better results when models were tested on the datasets they were trained on compared to out-of-domain datasets. Key evaluation metrics like MAP, MAR, and MAF1 were used to assess model performance in multimodal image retrieval tasks. For example, ResNet50 showed a MAP increase from 0.70 to 0.80 on the MSCOCO dataset under base loss conditions. Augmentation techniques further improved performance, with a MAP of 0.82 on the Flickr30k dataset. EfficientNet0 achieved a MAP of 0.84 on Flickr30k, improving to 0.87 with data augmentation. The Vision Transformer and Tiny Swin Transformer also demonstrated strong performance, with MAPs of 0.80 and 0.84 on Flickr8k, respectively. These findings highlight the significant impact of architecture variation and data augmentation on model generalization across domains, showing that careful data enhancement and loss function optimization can balance performance and computational efficiency.

\section{Limitations}

Our study on multimodal vision-language retrieval systems for low-resource languages has identified key limitations:

\noindent
\textbf{Synthetic Data Quality}: The performance heavily depends on the accuracy of synthetic data, such as translated captions from English to Azerbaijani. Ensuring high-quality data, potentially through human validation, is critical to avoid inaccuracies and biases.

\noindent
\textbf{Scalability and Generalization}: While transitioning from one language to another can be made relatively smooth, moving from one domain to another within the same language remains a significant challenge. Models trained in one domain often struggle to generalize across others without extensive tuning.

\section{Conclusion and Future Work}

This work developed an application of multimodal vision-language models for image retrieval in low-resource languages, with a focus on Azerbaijani. By using synthetic datasets—translated sentences from English to Azerbaijani via machine translation and data augmentation—the research demonstrated improved retrieval accuracy over traditional models. Key future directions include:

\begin{itemize}
\item \textbf{Expansion to Additional Low-Resource Languages and Enhanced Data Augmentation:} Extending the approach to other low-resource languages, incorporating advanced data augmentation, and involving human review to improve synthetic dataset quality could enhance model performance.

\item \textbf{Evaluation of Architectures Across Domains:} Testing various architectures on the translated ROCO dataset, especially in diverse domains like medical imaging, could offer valuable insights.

\item \textbf{Image Captioning and Energy Efficiency Optimization:} Exploring the reverse task of image captioning and optimizing model efficiency for resource-limited settings could enable deployment on mobile devices.

\item \textbf{Interactive User Feedback Systems:} Developing systems that incorporate user feedback to refine model outputs could improve retrieval accuracy and user satisfaction.

\end{itemize}

Further research in these areas will enhance the accessibility and applicability of AI technologies globally, with the ongoing release of configurations and collaborative advancements.
\newpage
\bibliography{custom}
\appendix{}
\section{Implementation}

\begin{algorithm}[H]
\caption{Dataset Preparation}
\begin{algorithmic}[1]
\Require $df, img\_dir, tokenizer, transform$
\Procedure{CLIPDataset}{$df, img\_dir, tokenizer, transform$}
    \State $self.dataframe \gets dataframe$
    \State $self.img\_dir \gets img\_dir$
    \State $self.tokenizer \gets tokenizer$
    \State $self.transform \gets transform$
\EndProcedure
\end{algorithmic}
\end{algorithm}

\begin{algorithm}[H]
\caption{Model Initialization}
\begin{algorithmic}[1]
\Procedure{ImageEncoder}{type, pretrained}
    \State $encoder\_type \gets type$
    \If{$encoder\_type = \text{'resnet50'}$}
        \State Initialize ResNet50
    \ElsIf{$encoder\_type = \text{'efficientnet'}$}
        \State Initialize EfficientNet-B0
    \ElsIf{$encoder\_type = \text{'vit'}$}
        \State Initialize Vision Transformer (ViT)
    \ElsIf{$encoder\_type = \text{'swin\_transformer'}$}
        \State Initialize Swin Transformer
    \ElsIf{$encoder\_type = \text{'convnext'}$}
        \State Initialize ConvNeXt
    \EndIf
    \State Load pretrained weights if applicable
\EndProcedure

\Procedure{TextEncoder}{model\_name, pretrained}
    \State Initialize text encoder (BERT-like)
    \State Load pretrained weights if applicable
\EndProcedure

\Procedure{CLIPModel}{image\_encoder, text\_encoder, output\_dim}
    \State Initialize encoders
    \State Attach projection heads
\EndProcedure
\end{algorithmic}
\end{algorithm}

\begin{figure}[H]
    \centering
    \includegraphics[width=0.6\linewidth]{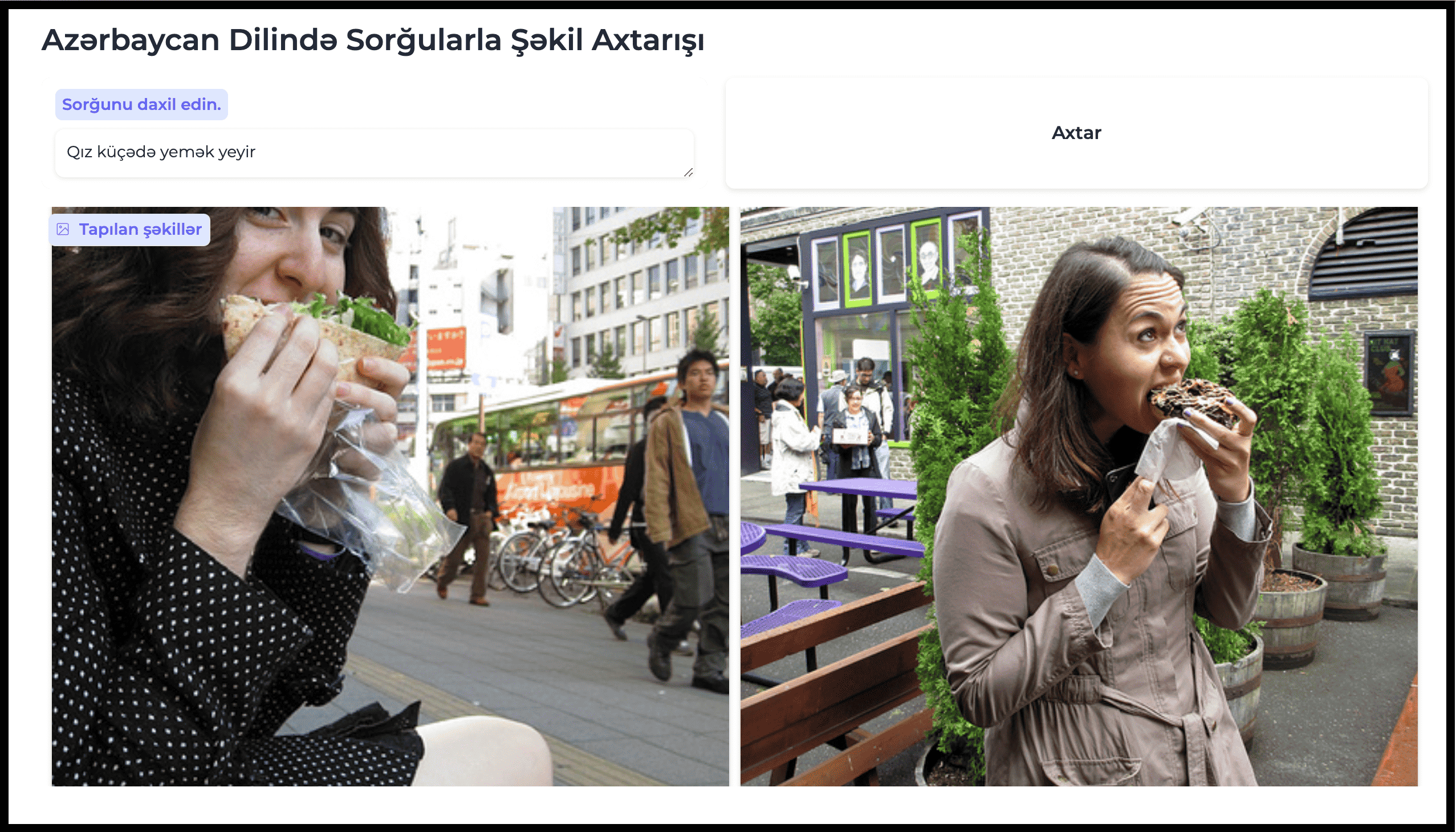}
    \caption{\textbf{Model Inference Example 2.} The query posed is "A girl eating on the street".}
    \label{fig:sample_3}
\end{figure}

\begin{algorithm}[H]
\caption{Training Loop}
\begin{algorithmic}[1]
\Require $train\_loader, model, optimizer, scheduler$
\State Initialize GPU setup (e.g., Nvidia T4)
\While{not converged}
    \For{images, input\_ids, attention\_masks in train\_loader}
        \State Forward pass with $images, input\_ids, attention\_masks$
        \State Calculate contrastive loss
        \State Backpropagate loss and update model parameters
        \State Update learning rate with scheduler
    \EndFor
\EndWhile
% \State Save model for evaluation
\end{algorithmic}
\end{algorithm}
\label{sec:appendix}

\noindent
\begin{table}[ht]
\centering
\begin{tabularx}{\linewidth}{|c|>{\RaggedRight\arraybackslash}X|>{\RaggedRight\arraybackslash}X|}
    \hline
    \textbf{Image ID} & \textbf{English Caption} & \textbf{Azerbaijani Translation} \\
    \hline
    203564 & A bicycle replica with a clock as the front wheel. & Ön teker kimi saat olan velosiped nüsxesi. \\
    \hline
    322141 & A room with blue walls and a white sink and door. & Mavi divarları, ağ lavabo ve qapısı olan otaq. \\
    \hline
    16977 & A car that seems to be parked illegally behind a legally parked car. & Qanuni park edilmiş avtomobilin arxasında qeyri-qanuni dayanmış kimi görünen avtomobil. \\
    \hline
    106140 & A large passenger airplane flying through the air. & Havada uçan böyük bir sernişin teyyaresi. \\
    \hline
\end{tabularx}
\caption{Sample from the training dataset.}
\label{tab:example_table}
\end{table}

\begin{figure}
    \centering
    \includegraphics[width=0.7\linewidth]{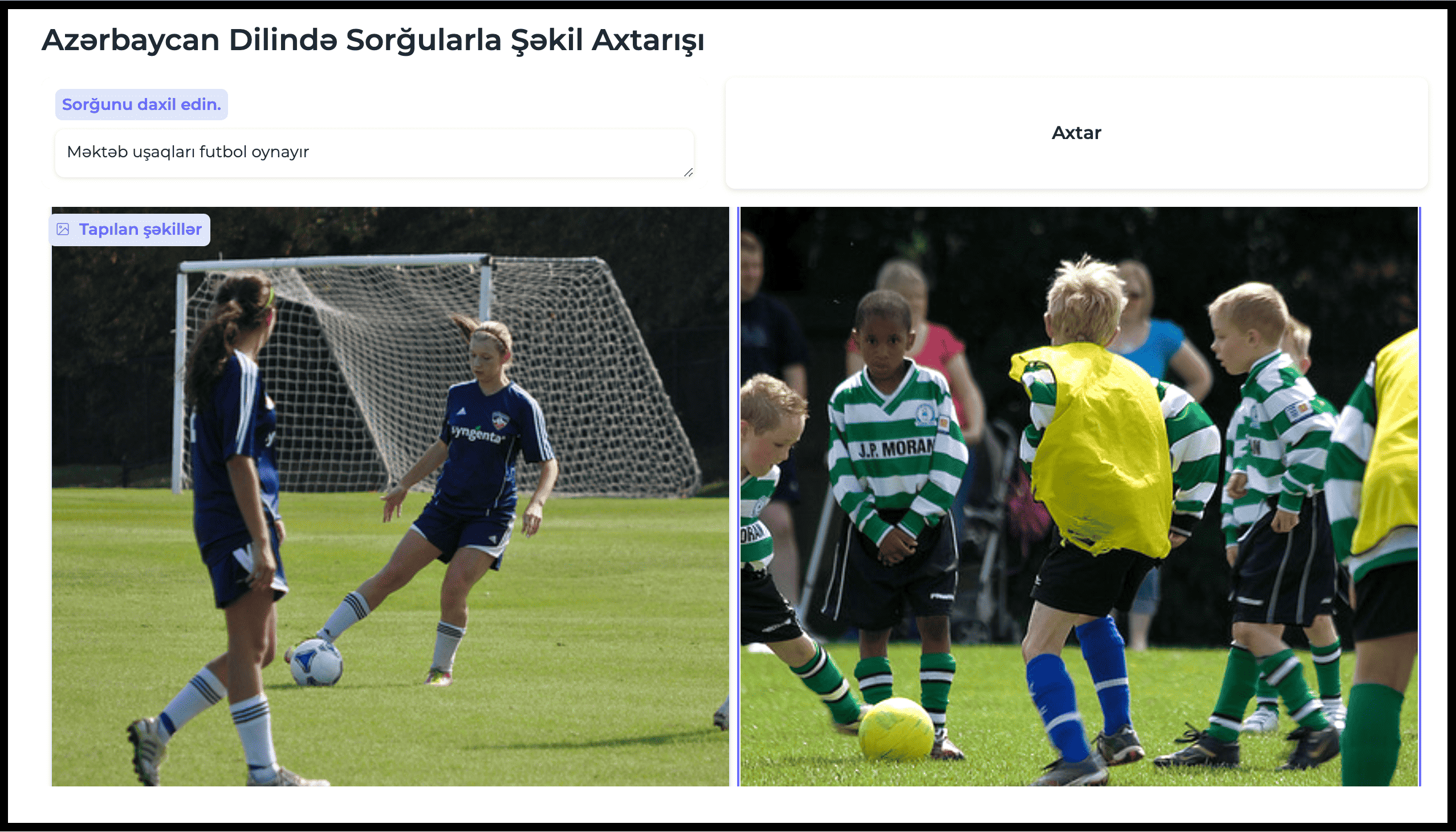}
    \caption{\textbf{Model Inference Example 3.} The query posed is "School children are playing football".}
    \label{fig:sample_4}
\end{figure}

\end{document}